\documentclass[10pt,twocolumn,letterpaper]{article}

\usepackage{cvpr}

\usepackage[numbers]{natbib}
\usepackage{subcaption}
\usepackage{makecell}
\usepackage{amsmath,amssymb,amsfonts}
\usepackage{algorithmic}
\usepackage{graphicx}
\usepackage{textcomp}
\usepackage{booktabs}
\usepackage{multirow}
\usepackage{xspace}
\usepackage{float}
\usepackage{icomma}
\usepackage[x11names,table]{xcolor}

\definecolor{cvprblue}{rgb}{0.21,0.49,0.74}
\usepackage[pagebackref,breaklinks,colorlinks,allcolors=cvprblue]{hyperref}

\def\paperID{10}
\def\confName{CVPR}
\def\confYear{2026}

\title{\Large \bf
\textsc{DriveXQA}: Cross-modal Visual Question Answering for Adverse Driving Scene Understanding
}

\author{Mingzhe Tao$^{1}$\thanks{Equal contribution.}, Ruiping Liu$^{1,*}$\thanks{Project lead.}, Junwei Zheng$^{1}$, Yufan Chen$^{1}$, Kedi Ying$^{1}$, M. Saquib Sarfraz$^{1,3}$,\\Kailun Yang$^{2}$, Jiaming Zhang$^{2}$\thanks{Corresponding author.}, and Rainer Stiefelhagen$^{1}$
  \\[0.4em]
  \small $^{1}$Karlsruhe Institute of Technology \quad $^{2}$Hunan University \quad $^{3}$Mercedes-Benz Tech Innovation
}

\begin{document}
\maketitle
\begin{abstract}
Fusing sensors with complementary modalities is crucial for maintaining a stable and comprehensive understanding of abnormal driving scenes. However, Multimodal Large Language Models (MLLMs) are underexplored for leveraging multi-sensor information to understand adverse driving scenarios in autonomous vehicles. To address this gap, we propose the \textsc{DriveXQA}, a multimodal dataset for autonomous driving VQA. In addition to four visual modalities, five sensor failure cases, and five weather conditions, it includes $102,505$ QA pairs categorized into three types: global scene level, allocentric level, and ego-vehicle centric level. Since no existing MLLM framework adopts multiple complementary visual modalities as input, we design \textsc{MVX-LLM}, a token-efficient architecture with a Dual Cross-Attention (DCA) projector that fuses the modalities to alleviate information redundancy. Experiments demonstrate that our DCA achieves improved performance under challenging conditions such as foggy (GPTScore: $53.5$ \textit{vs.} $25.1$ for the baseline). The dataset and source code are at \url{https://github.com/jtjmd/DRIVEXQA}. 
\end{abstract}    
\section{Introduction}
\label{sec:intro}
\begin{figure}[t]
\centering
\includegraphics[width=0.98\columnwidth]{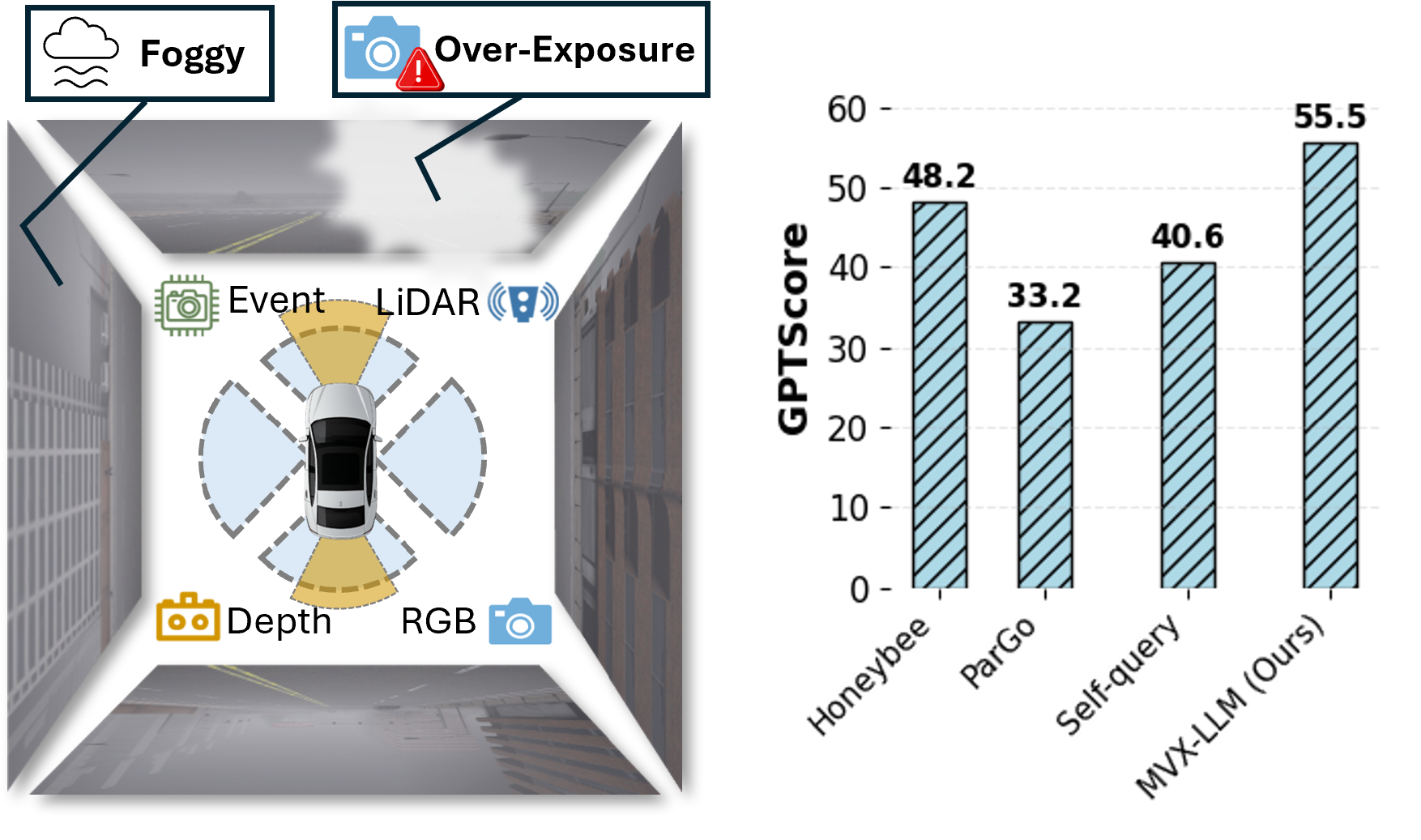}
\caption{\textbf{Left}: Two corner cases of adverse driving scenes (\textit{foggy} condition causing poor visibility and camera \textit{over-exposure} resulting in degraded image quality). \textbf{Right}: Performance (GPTScore) comparison on \textsc{DriveXQA} dataset.}
\label{fig:intro}
\end{figure}

Autonomous vehicles require a holistic understanding of complex driving environments through multiple sensor modalities to ensure safe navigation. While existing MLLMs have achieved remarkable success in general vision-language tasks~\cite{liu2023llava, alayrac2022flamingo}, they remain underexplored for leveraging multi-sensor information in safety-critical autonomous driving scenarios. Unlike general visual understanding tasks, driving scenarios demand a precise understanding of spatial relationships, environmental conditions, and sensor reliability under adverse conditions such as foggy, rainy, and various sensor failures. As illustrated in Fig.~\ref{fig:intro}, adverse driving conditions such as \textit{foggy} and camera \textit{over-exposure} represent critical challenges that existing systems fail to address comprehensively. These abnormal scenarios are commonplace in real-world deployment yet severely compromise perception quality and system reliability. 

Visual Question Answering (VQA) systems serve as critical auxiliary cognition components in autonomous driving, enabling vehicles to understand and reason about complex driving scenarios through natural language interactions. However, current autonomous driving VQA systems primarily focus on normal conditions with limited sensor diversity and inadequate handling of abnormal situations. Existing datasets, including NuScenes-QA~\cite{qian2024nuscenes}, DriveLM~\cite{sima2024drivelm}, and LingoQA~\cite{marcu2024lingoqa}, lack systematic coverage of adverse environmental conditions and multi-modal sensor fusion strategies under sensor degradation scenarios.
MLLMs~\cite{yin2024survey,caffagni2024revolution} as auxiliary cognition systems in autonomous driving should ideally leverage multiple complementary sensor modalities to adapt to adverse conditions through cross-modal compensation, where degraded information from one sensor can be compensated by reliable signals from others. while recent VLA models and BEV-based Transformers fuse multi-sensor data for action prediction, no existing MLLM-based VQA framework simultaneously supports multiple complementary visual modalities with token efficiency and robustness under sensor degradation. Current approaches~\cite{tziafas2023early,zhang2024rethinking} typically rely on single-modality inputs (predominantly RGB) or simplistic fusion mechanisms that fail to leverage the complementary nature of different sensor modalities effectively. This limitation becomes particularly problematic under adverse conditions where individual sensors may provide degraded or unreliable information, when the interpretable reasoning capabilities of VQA systems are most needed to support critical driving decisions. For instance, in heavy fog or nighttime driving, cameras lose visibility and LiDAR returns become unreliable, exactly the scenarios where a driver most urgently needs the system to explain what it sees and why it recommends a particular action.

To address these limitations, we first propose the \textsc{\textbf{DriveXQA}} dataset for evaluating MLLMs in adverse driving scene understanding. \textsc{DriveXQA} contains $7,885$ frames of driving scenes, systematically covering diverse weather conditions including \textit{rainy}, \textit{night}, \textit{foggy}, \textit{cloudy}, and \textit{sunny} scenarios, along with five types of sensor failures: \textit{motion blur} (MB), \textit{overexposure} (OE), \textit{underexposure} (UE), \textit{LiDAR jitter} (LJ), and \textit{event low-resolution} (EL).
Second, we propose \textsc{MVX-LLM}, a token-efficient architecture that fuses multiple sensor modalities from \textbf{Multiple Views} effectively with a \textbf{Dual Cross-Attention (DCA)} projector. This architecture is designed to address abnormal conditions in autonomous driving scenarios through integrated abnormal condition modeling and cross-modal vision-language fusion. As demonstrated in Figure~\ref{fig:intro}, our MVX-LLM achieves the highest GPT score of $55.5$, significantly outperforming baseline methods without injected adverse condition knowledge. While Honeybee and ParGo~\cite{cha2024honeybee,wang2025pargo} represent conventional projector approaches that are not effective to deal with multiple modalities, and Self-query~\cite{zhang2023delivering} incorporates abnormal handling but lacks cross-modal vision-language integration, our approach combines both capabilities of dealing with multiple modalities and comprehensive cross-modal vision-language integration.
\begin{table*}[!t]
    \vspace{2mm}
    \centering
    \renewcommand{\arraystretch}{0.55}
    \setlength{\tabcolsep}{8mm}
    \caption{Comparison of autonomous driving VQA datasets. \textbf{MV}: Multi-View RGB cameras.  \textbf{SF}: Sensor Failure scenarios. \#VQA: Number of VQA pairs. \textbf{FT}: Fine-tuned Evaluation. \textbf{Hierarchical}: Hierarchical question structure. $\triangle$ means partially involved.}
    \label{tab:dataset_comparison}
    \resizebox{\textwidth}{!}{
    \begin{tabular}{lccccccc}
        \toprule
        \textbf{Dataset} & \textbf{multi-view} & \textbf{modalities} & \textbf{SF} & \textbf{\#VQA} & \textbf{FT} & \textbf{Hierarchical} \\
        \midrule
        DRAMA~\cite{malla2023drama} & $\times$ & 1 & $\times$ & 17K & $\checkmark$ & $\times$ \\
        VLAAD~\cite{park2024vlaad} & $\times$ & 1 & $\times$ & 64K & $\checkmark$ & $\times$ \\
        LingoQA~\cite{marcu2024lingoqa} & $\times$ & 1 & $\times$ & 419K & $\checkmark$ & $\times$ \\
        NuPlanQA~\cite{park2025nuplanqa} & $\checkmark$ & 1 & $\times$ & 1M & $\checkmark$ & $\triangle$ \\
        NuScenes-QA~\cite{qian2024nuscenes} & $\checkmark$ & 2 & $\times$ & 460K & $\checkmark$ & $\times$ \\
        NuScenes-MQA~\cite{inoue2024nuscenes} & $\checkmark$ & 2 & $\times$ & -- & $\checkmark$ & $\times$ \\
        DriveLM~\cite{sima2024drivelm} & $\checkmark$ & 2 & $\times$ & -- & $\times$ & $\triangle$ \\
        \rowcolor[gray]{.9} \textbf{\textsc{DriveXQA} (Ours)} & $\checkmark$ & 4 & $\checkmark$ & 102K & $\checkmark$ & $\checkmark$ \\
        \bottomrule
    \end{tabular}}
\end{table*}
The main contributions of this work are:
\begin{itemize}
\item A comprehensive cross-modal driving VQA dataset \textsc{DriveXQA} considering adverse conditions including diverse weather variations and sensor failure scenarios, with a systematic hierarchical QA taxonomy covering global scene, allocentric, and ego-vehicle centric levels.
\item A token-efficient multi-modal architecture \textsc{MVX-LLM} for robust sensor fusion under adverse conditions.
\item Extensive evaluation demonstrating effectiveness under challenging weather conditions and sensor degradation scenarios.
\end{itemize}

\section{Related Work}
\subsection{Autonomous Driving VQA Datasets and Benchmarks}
Visual Question Answering (VQA) is a task that enables machines to comprehend visual content and provide natural language responses to structured queries. In the context of autonomous driving, VQA systems primarily serve as auxiliary reasoning components that support onboard decision-making by translating complex perceptual inputs into interpretable natural language representations, facilitating both system diagnostics and human oversight. Previous VQA datasets~\cite{antol2015vqa, goyal2017making, johnson2017clevr, hudson2019gqa} established foundational evaluation principles for visual reasoning, introducing compositional question structures and systematic evaluation of spatial and relational understanding. Building on these foundations, driving-specific VQA datasets have progressively expanded in scope and modality coverage. NuScenes-QA~\cite{qian2024nuscenes} pioneered driving-specific VQA with multi-modal inputs, including RGB and LiDAR for dynamic outdoor environments. NuScenes-MQA~\cite{inoue2024nuscenes} introduced MarkupQA for unified evaluation, DriveLM~\cite{sima2024drivelm} combined graph-based VQA for spatial reasoning, and MapLM~\cite{cao2024maplm} addressed traffic scene evaluation, with further specialized work on uncertainty~\cite{ouyang2022autonomous}, object-level reasoning~\cite{chen2024driving}, and 3D comprehension~\cite{zhu2024llava, choudhary2024talk2bev}. NuPlanQA~\cite{park2025nuplanqa} presented large-scale multi-view driving scene understanding across multiple subtasks (\textit{e.g.}, traffic light detection, spatial relations recognition, ego-vehicle maneuver prediction), but lacks depth cameras and systematic sensor failure coverage. LingoQA~\cite{marcu2024lingoqa} introduced video-based driving VQA with temporal reasoning, but focuses exclusively on RGB inputs without complementary sensor modalities. Recent specialized datasets, including DRAMA~\cite{malla2023drama}, VLAAD~\cite{park2024vlaad}, and additional work~\cite{chiu2025v2v}, have addressed hazard prediction, interaction reasoning, and cooperative driving.

Despite this progress, existing datasets suffer from significant limitations that hinder robust evaluation under real-world deployment conditions. VQA model performance evaluation studies~\cite{rekanar2023towards, rekanar2024subjective} identified limitations including insufficient spatial reasoning and environmental robustness, and multi-modal datasets~\cite{zhang2023delivering, caesar2020nuscenes, sun2020scalability} only partially address the need for rigorous sensor fusion evaluation. As summarized in Table~\ref{tab:dataset_comparison}, these works share four critical gaps: (1) lack of comprehensive multi-modal sensor coverage, particularly event cameras; (2) absence of sensor failure scenarios; (3) insufficient hierarchical question structures; and (4) limited scale for training and evaluation. Our \textsc{DriveXQA} is specifically designed to address all of these limitations.
\subsection{Multi-Modal Large Language Models for Sensor Fusion}
The emergence of large language models has fundamentally transformed multi-modal understanding, with foundational works~\cite{chen2024sharegpt4v, liu2023llava, li2023blip, alayrac2022flamingo} establishing vision-language integration paradigms. LLaVA~\cite{liu2023llava} pioneered two-stage training with feature alignment pretraining and instruction tuning.
Within vision-language models, efficient visual token management became critical. Honeybee~\cite{cha2024honeybee} introduces locality-enhanced projectors for computational efficiency while preserving spatial context. ParGo~\cite{wang2025pargo} advances this with Partial-Global projectors integrating both global and partial visual representations to preserve fine-grained details while mitigating overemphasis on prominent regions. Parameter-efficient methods such as LoRA~\cite{hu2022lora}, QLoRA~\cite{dettmers2023qlora}, and related strategies~\cite{peft} enable practical fine-tuning of large-scale models under constrained budgets.

However, real-world autonomous driving requires multiple complementary sensors beyond vision-language pairs. Cross-modal alignment research emerged to handle semantic correspondence across sensor modalities through contrastive learning~\cite{dufumier2024align, zhang2024multi, lin2022multi} and multi-scale integration~\cite{cicchetti2024gramian}, enabling consistent alignment across diverse spatial and temporal characteristics.
Dynamic environmental conditions further complicate multi-sensor fusion. Condition-aware fusion strategies~\cite{brodermann2025cafuser, wang2022multimodal, yin2023dformer, yin2025dformerv2} introduced adaptive mechanisms that adjust fusion weights based on environmental context and sensor reliability. 
Despite these advances, existing multi-modal LLM frameworks fundamentally cannot accept multiple complementary sensor modalities as input, making them inadequate for autonomous driving scenarios requiring robust performance under sensor failures and adverse conditions. Our work addresses this limitation by proposing \textsc{MVX-LLM}, a specialized architecture that fuses multiple sensor inputs for robust multi-sensor autonomous driving applications.
\section{\textsc{DriveXQA} Dataset}
\subsection{Dataset Generation}
Our \textsc{DriveXQA} dataset comprises $7,885$ frames of driving scenes with $102,505$ question-answer pairs, following a systematic $80\%/10\%/10\%$ train/validation/test split. Scenes are collected in CARLA (v0.9.14)~\cite{dosovitskiy2017carlaopenurbandriving} and sensor failures are injected via post-processing. QA pairs are generated by prompting GPT-4o with structured scene-metadata templates. Manual quality control covering a stratified $5\%$ sample was then conducted to ensure answer quality, correcting weather misclassification and inaccurate counts, and validating ego-centric questions by cross-referencing LiDAR and multi-view RGB data.
$13$ QA pairs are generated for each driving scene according to our hierarchical framework ($2$ global scene level, $8$ allocentric level, $3$ ego-vehicle centric level), ensuring proportional representation across question categories. The question-answer pairs demonstrate balanced linguistic complexity with an average question length of $11.4$ words and answer length of $12.9$ words, reflecting concise yet information-dense annotations that support reliable automatic evaluation across diverse question types and difficulty levels.
\begin{figure}[t]
    \centering
    \includegraphics[width=0.94\linewidth]{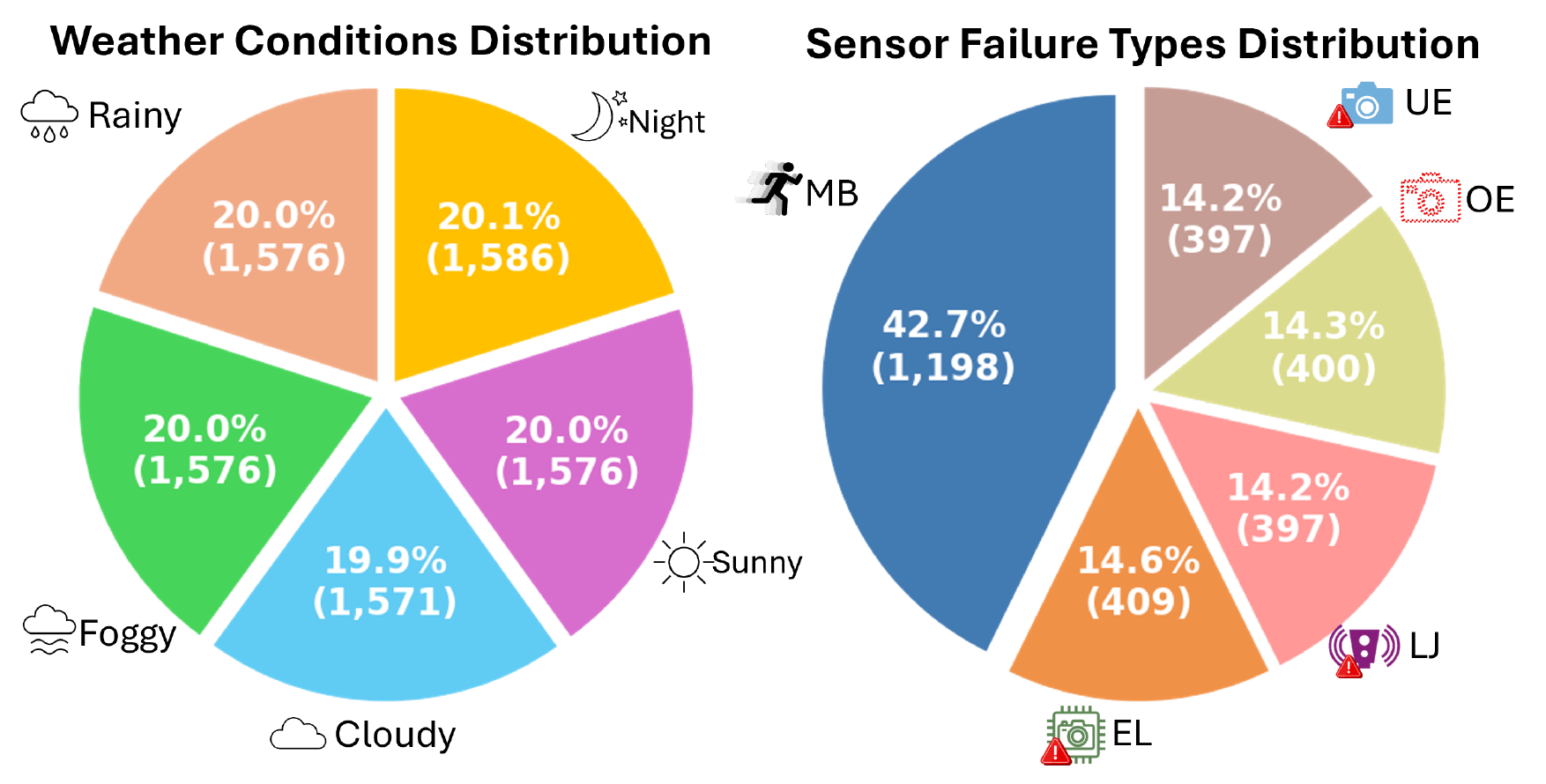}
    \caption{Statistics of \textsc{DriveXQA} dataset showing distribution of weather conditions (left) and distribution of sensor failure types (right): MB (Motion Blur), OE (Overexposure), UE (Underexposure), LJ (LiDAR Jitter), EL
(Event Low-resolution).}
    \label{fig:dataset_statistics}
\end{figure}
Fig.~\ref{fig:dataset_statistics} presents the broad coverage of our dataset across environmental conditions and sensor degradation scenarios. The balanced weather distribution covers all realistic automotive scenarios, while the systematic inclusion of sensor failures enables evaluation of system resilience under realistic implementation cases with $35.5\%$ corner cases.
The dataset covers balanced weather conditions across Rainy ($20.0\%$), Night ($20.1\%$), Sunny ($20.0\%$), Cloudy ($19.9\%$), and Foggy ($20.0\%$). Sensor failures include Motion Blur ($42.7\%$), Underexposure ($14.2\%$), Overexposure ($14.3\%$), LiDAR Jitter ($14.2\%$), and Event Low-resolution ($14.6\%$), representing $35.5\%$ of total samples as corner cases.
\begin{figure}[t]
    \centering
    \includegraphics[width=0.99\linewidth]{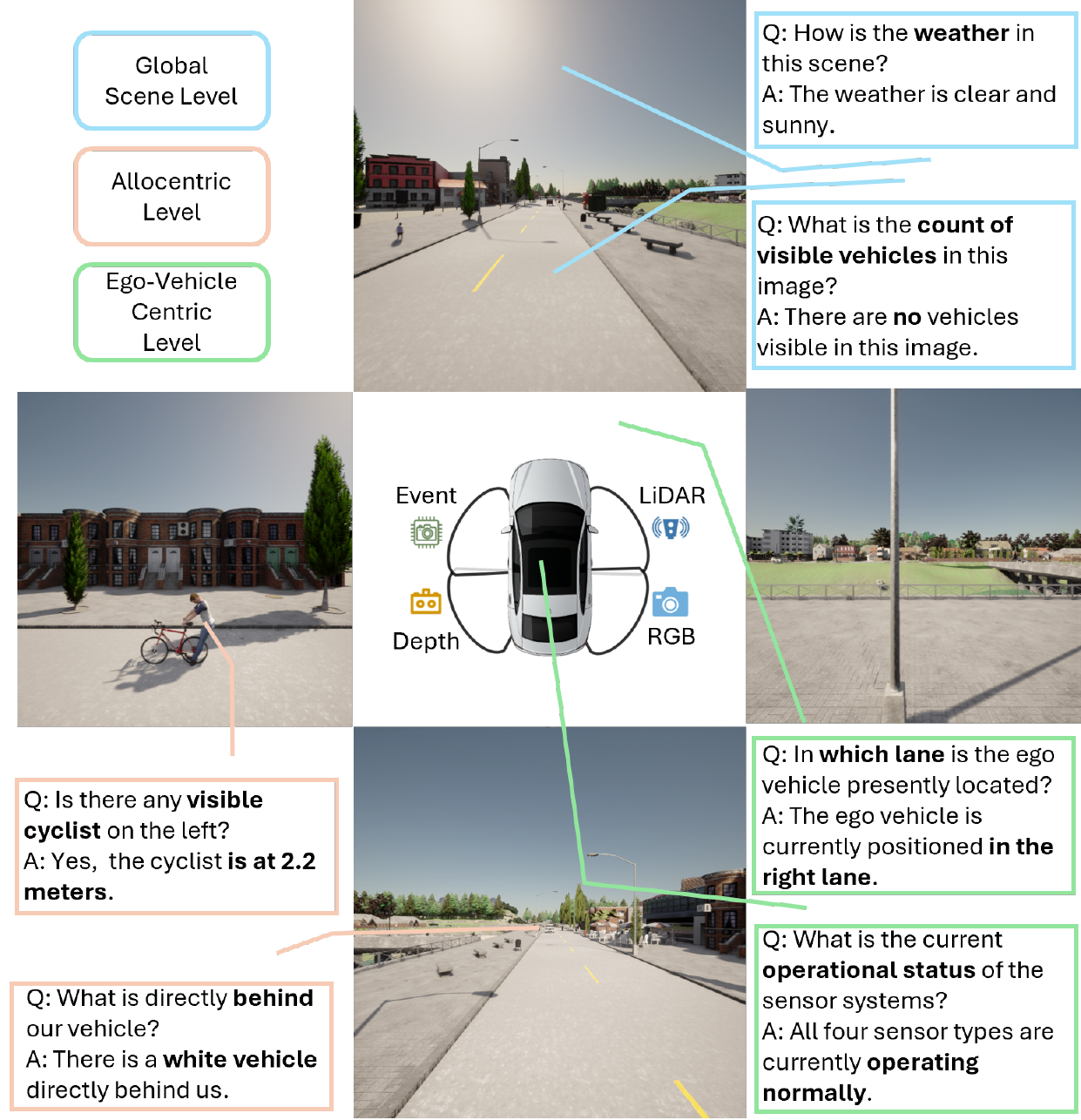}
    \caption{Hierarchical XQA examples on \textsc{DriveXQA} dataset. The framework demonstrates three semantic levels: \textit{Global Scene Level}, \textit{Allocentric Level}, and \textit{Ego-Vehicle Centric Level}.
    }
    \label{fig:qa_examples}
\end{figure}
\subsection{Hierarchical XQA}
We develop a comprehensive XQA dataset structured across three receptive fields, \textit{i.e.} global scene level, allocentric level, and ego-vehicle centric level. 
Exact $13$ questions are generated from each driving scene following our hierarchical framework (2 global scene level, 8 allocentric level, 3 ego-vehicle centric level).
\subsubsection{Global Scene Level}
Global scene questions assess overall environmental conditions and traffic patterns that cannot be reliably inferred from any single sensor modality alone. These questions target weather-level perception (\textit{e.g.}, visibility estimation, precipitation recognition) and macro-level traffic state assessment, requiring holistic fusion of multi-modal inputs to reason about scene-wide factors that directly influence driving strategy and system reliability. These questions focus on weather impact and traffic density, providing essential context for the decision-making of autonomous driving. This level considers broad environmental factors that influence overall driving strategy and system reliability.
\subsubsection{Allocentric Level}
Allocentric questions cover spatial relationships in the scene through structured queries, including quantitative spatial analysis, distance measurements, traffic sign interpretation, road condition assessment, and object categorization. 
This level leverages multi-view visual information from four camera perspectives (front, back, left, right) to enable comprehensive spatial reasoning and cross-view consistency validation (\textit{e.g.}, \textit{``What is the distance to the cyclist on the left?''}). Multi-view perspectives are crucial at this level because they offer complete scene coverage, enabling both inter-object relationship reasoning and spatial awareness for navigation planning.
\begin{figure*}[t]
    \vspace{2mm}
    \centering
    \includegraphics[width=0.95\textwidth]{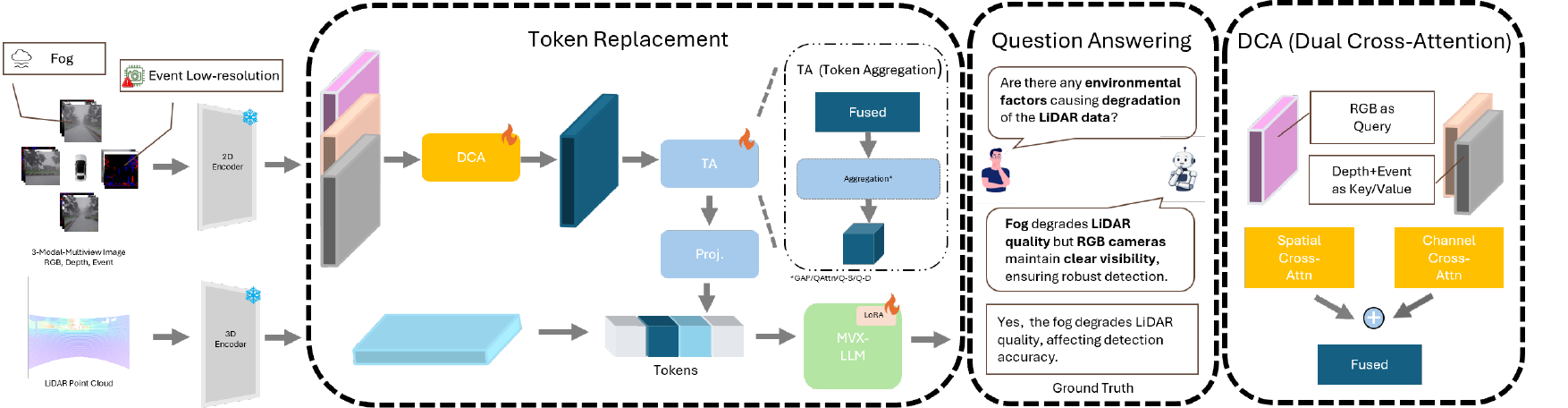}
    \caption{Overview of \textsc{MVX-LLM}. 
    The framework processes multi-modal sensor inputs (RGB, Depth, Event cameras from four viewpoints, and LiDAR point clouds) through specialized encoders. The DCA mechanism integrates RGB, Depth, and Event features before token replacement. The Question Answering component utilizes the fused representations to generate hierarchical responses across Global Scene, Allocentric, and Ego-Vehicle Centric levels under adverse driving conditions.}
    \label{fig:system_overview}
\end{figure*}
\subsubsection{Ego-Vehicle Centric Level}
Egocentric questions focus on direct environment perception and vehicle state, including lane positioning analysis, sensor health monitoring, and surrounding vehicle behavior prediction. These queries address safety-critical aspects requiring a precise understanding of the ego-vehicle's operational environment and immediate decision-making requirements.

Fig.~\ref{fig:qa_examples} demonstrates our hierarchical QA structure through concrete examples. The global level question assesses weather impact on visibility, the allocentric level analyzes traffic density and spatial relationships, while the egocentric level focuses on lane positioning and operational status. This systematic approach ensures a comprehensive evaluation of autonomous driving capabilities across different analysis perspectives and environmental conditions.
\section{\textsc{MVX-LLM} Framework}
\subsection{Overall Architecture}
To solve the XQA task, we propose the \textsc{MVX-LLM} framework (Fig.~\ref{fig:system_overview}) that processes multi-modal sensor inputs through specialized encoders: a shared CLIP-based~\cite{radford2021learning} Vision Transformer for RGB, depth, and event camera data from four perspectives, and a PointNet++~\cite{qi2017pointnet++} encoder for LiDAR point clouds. 
Due to the heterogeneous modalities, LiDAR data is processed separately through hierarchical set abstraction operations and projected to unified $512$-dimensional representations without involving in the cross-modal fusion. Our framework employs a Dual Cross-Attention (DCA) mechanism for 2D modalities, specifically designed for robust multi-modal fusion when individual sensors experience degradation due to adverse weather or hardware failures.
\subsection{Dual Cross-Attention Mechanism}
Autonomous driving scenarios demand adaptive cross-modal fusion where individual sensors may experience degradation due to adverse weather or sensor failures. Traditional concatenation-based approaches~\cite{huang2024embodied} treat all modalities equally, lacking the ability to adapt to sensor failures. DCA mechanism addresses this limitation by operating through complementary spatial and channel cross-attention pathways that enable adaptive feature weighting based on environmental conditions and sensor availability.
All modalities undergo spatial alignment from a native $49$-token representation (derived from the $7×7$ spatial grid output of the CLIP-based ViT encoder) to a $48$-token configuration for computational efficiency and attention head compatibility:
\begin{align}
\mathbf{F}_{rgb}^{'} &= \mathbf{F}_{rgb} \mathbf{W}_{align}^{rgb} \in \mathbb{R}^{B \times 48 \times 512} \\
\mathbf{F}_{depth}^{'} &= \mathbf{F}_{depth} \mathbf{W}_{align}^{depth} \in \mathbb{R}^{B \times 48 \times 512} \\
\mathbf{F}_{event}^{'} &= \mathbf{F}_{event} \mathbf{W}_{align}^{event} \in \mathbb{R}^{B \times 48 \times 512}
\end{align}
The spatial cross-attention branch establishes spatial correspondences between modalities, enabling cross-modal information exchange at specific spatial locations:
\begin{equation}
\mathbf{F}_{s} = \text{MultiHeadAttn}_{s}(\mathbf{Q}_{rgb}, \mathbf{K}_{multi}, \mathbf{V}_{multi})
\end{equation}
where $\mathbf{Q}_{rgb} = \mathbf{F}_{rgb}^{aligned} \in \mathbb{R}^{B \times 48 \times 512}$ and:
\begin{equation}
\mathbf{K}_{multi} = \mathbf{V}_{multi} = [\mathbf{F}_{depth}^{'}; \mathbf{F}_{event}^{'}] \in \mathbb{R}^{B \times 96 \times 512}
\end{equation}
The channel cross-attention branch operates on transposed features to enable cross-channel feature enhancement. Concatenated multi-modal features require alignment to resolve dimensional incompatibility with RGB queries:
\begin{equation}
\mathbf{F}_{c} = \text{MultiHeadAttn}_{c}(\mathbf{Q}_{rgb}^{\top}, \mathbf{K}_{multi}^{'}, \mathbf{V}_{multi}^{'})
\end{equation}
where:
\begin{align}
\mathbf{Q}_{rgb}^{\top} &= (\mathbf{F}_{rgb}^{'})^{\mathrm{T}} \in \mathbb{R}^{B \times 512 \times 48} \\
\mathbf{K}_{multi}^{'} = \mathbf{V}_{multi}^{'} &= [\mathbf{F}_{depth}^{'}; \mathbf{F}_{event}^{'}]^{\mathrm{T}} \mathbf{W}_{align}
\end{align}
with learned transformation $\mathbf{W}_{align} \in \mathbb{R}^{96 \times 48}$ ensuring $\mathbf{K}_{multi}^{’} \in \mathbb{R}^{B \times 512 \times 48}$.
The final fused representation combines both pathways:
\begin{equation}
\mathbf{F}_{fused} = \frac{\mathbf{F}_{s}+\mathbf{F}_{c}^{\mathrm{T}}}{2}
\end{equation}
This dual-pathway design captures both spatial-level and channel-level cross-modal interactions, providing a comprehensive multi-modal fusion capability.
\subsection{Token Aggregation}
The token aggregation module processes the fused multi-modal features through specialized fusion heads that compress the visual tokens from multi-modalities into compact token embeddings suitable for language model integration. Following TokenFusion~\cite{wang2022multimodal} principles, we compare four distinct aggregation strategies to address different aspects of spatial feature consolidation while maintaining computational efficiency.
\subsubsection{Global Average Pooling (GAP)} 
The Global Average Pooling head provides the baseline aggregation mechanism through simple mean pooling across all spatial tokens: 
\begin{equation} 
\mathbf{F}_{gap} = \frac{1}{N} \sum_{i=1}^N \mathbf{F}_{fused}[i,:] \in \mathbb{R}^{B \times 512}
\end{equation} 
where $N=48$ represents the number of spatial tokens. GAP serves as the computational efficiency baseline, requiring no additional parameters.
\subsubsection{Query Attention (QAttn)} 
The Query Attention mechanism implements learnable query-based aggregation through multi-head cross-attention: 
\begin{equation} 
\mathbf{F}_{qattn} = \text{MultiHeadAttn}(\mathbf{Q}_{learn}, \mathbf{F}_{fused}, \mathbf{F}_{fused})
\end{equation} 
where $\mathbf{Q}_{learn} \in \mathbb{R}^{1 \times 512}$ represents learnable query. The learnable queries enable the model to focus on task-relevant spatial regions during training, significantly improving performance over uniform pooling strategies.
\subsubsection{Spectral QAttn} 
The Spectral QAttn variant incorporates frequency domain processing through depthwise convolution before query attention: 
\begin{equation} 
\mathbf{F}_{enhanced} = [\mathbf{F}_{fused}; \text{DWConv1D}(\mathbf{F}_{fused})]
\end{equation} 
\begin{equation} 
\mathbf{F}_{spectral} = \text{MultiHeadAttn}(\mathbf{Q}_{learn}, \mathbf{F}_{enhanced}, \mathbf{F}_{enhanced})
\end{equation}
\subsubsection{DepthGate QAttn} 
The DepthGate QAttn variant incorporates depth-guided confidence estimation inspired by DFormerv2~\cite{yin2025dformerv2}: 
\begin{equation}
\mathbf{G}_{depth}^{(i)} = \sigma(\mathbf{F}_{depth}^{'}[i,:] \cdot \mathbf{w}_{gate} + b_{gate})
\end{equation}
\begin{equation}
\mathbf{F}_{depthgate} = \text{MultiHeadAttn}(\mathbf{Q}_{learn}, \mathbf{F}_{gated}, \mathbf{F}_{gated})
\end{equation}
Based on experimental validation, QAttn achieves the highest performance (GPTScore: $55.5$) and serves as our final implementation choice.
\section{Experiments}
\subsection{Implementation Details}
DCA mechanism operates with $8$ attention heads for spatial attention ($64$ dimensions each) and $4$ heads for channel attention ($12$ dimensions each), ensuring balanced computational load while maintaining full utilization of the $512$-dimensional feature space. Training is conducted using standard cross-entropy loss with Adam optimizer, learning rate of $1e{-}4$, and batch size of $16$. 
We employ LoRA fine-tuning with rank $16$ for parameter-efficient adaptation of the language model components. 
All experiments use $4$ NVIDIA A100-40GB GPUs with mixed-precision training. The DCA projector adds \textasciitilde$2.6M$ parameters ($<0.3\%$ of total model size), and per-modality token compression to a single query token keeps visual token count constant regardless of modality number, preserving inference efficiency.
For evaluation, we employ multiple metrics to assess different aspects of visual question answering performance. Following standard VQA evaluation protocols, we use BLEU-4, ROUGE-L, METEOR, and CIDEr for lexical and semantic similarity assessment. 
Additionally, we employ sentence embedding-based similarity scores. 
Following the LLM-as-a-judge paradigm~\cite{li2024llms, li2025preference}, we use GPT-4o-mini~\cite{hurst2024gpt} for semantic correctness evaluation, providing human-like judgment on factual accuracy and contextual appropriateness.
GPT-based evaluation uses a $1{\sim}5$ rubric normalized to $0{\sim}100$ via:
\begin{equation}
S
= \frac{1}{N_{samples}} \sum_{i=1}^{N_{samples}} \frac{s_i - 1}{4} \times 100\%,
\end{equation}
where $s_i$ represents the individual GPT score for sample $i$, and $N_{samples}$ is the total number of evaluated samples. 
To assess the reliability of GPT-based evaluation for DriveXQA, we recruited four human evaluators who independently assessed a subset of $360$ samples across different weather conditions and sensor failures. 
\subsection{Results on \textsc{DriveXQA}}
\begin{table*}[t]
\vspace{2mm}
\centering
\caption{Results on \textsc{DriveXQA} dataset across weather conditions and sensor failures. 
Each cell shows GPT-4o-mini score/Human evaluation score. {\color[HTML]{3531FF} \textbf{Blue}} indicates state-of-the-art results. Weather conditions: CL (Cloudy), FG (Foggy), NT (Night), RN (Rainy), SN (Sunny). Sensor failures: MB (Motion Blur), OE (Overexposure), UE (Underexposure), LJ (LiDAR Jitter), EL (Event Low-resolution).}
\label{tab:main_results}
\resizebox{\textwidth}{!}{%
\renewcommand{\arraystretch}{1.1}
\begin{tabular}{@{}lllllllllll|c@{}}
\toprule
\textbf{Method} & \textbf{CL} & \textbf{FG} & \textbf{NT} & \textbf{RN} & \textbf{SN} & \textbf{MB} & \textbf{OE} & \textbf{UE} & \textbf{LJ} & \textbf{EL} & \textbf{Average} \\ \midrule
prepend & {\color[HTML]{3531FF} \textbf{61.2}}/58.2 & 25.1/26.3 & {\color[HTML]{3531FF} \textbf{58.8}}/{\color[HTML]{3531FF} \textbf{60.1}} & 52.3/50.5 & 55.1/58.3 & 53.1/57.4 & 44.1/44.7 & 51.5/49.2 & {\color[HTML]{3531FF} \textbf{60.5}}/{\color[HTML]{3531FF} \textbf{62.5}} & {\color[HTML]{3531FF} \textbf{55.2}}/{\color[HTML]{3531FF} \textbf{58.1}} & 55.1/57.9 \\
 \midrule
\rowcolor[gray]{.9} \multicolumn{12}{l}{\textit{Token Fusion}} \\
Self-query & 40.9/43.4 & 40.3/42.7 & 34.9/37.6 & 41.1/43.5 & 48.1/46.3 & 39.9/36.7 & 28.9/28.4 & 36.1/39.8 & 39.5/42.5 & 44.1/47.6 & 40.6/43.3 \\
GAP & 51.2/54.5 & 53.2/56.4 & 50.5/53.6 & 50.2/47.4 & 50.4/53.8 & 54.2/57.5 & 49.4/52.7 & 51.1/48.4 & 54.9/57.2 & 54.9/57.3 & 50.5/53.1 \\
QAttn(Spectral) & 56.5/58.7 & 53.7/55.4 & 46.8/47.9 & 53.0/56.3 & 53.2/54.5 & 54.3/57.7 & 46.0/44.9 & 51.6/54.4 & 55.2/57.3 & 55.3/57.2 & 52.8/55.9 \\
QAttn(DepthGate) & 38.3/40.2 & 40.1/43.6 & 33.0/34.2 & 41.3/44.6 & 50.0/53.4 & 41.1/43.4 & 27.5/24.8 & 48.3/51.9 & 43.9/45.4 & 40.3/44.6 & 40.0/44.1 \\
 \midrule
\rowcolor[gray]{.9} \multicolumn{12}{l}{\textit{Projector Architectures}} \\
GAP + Honeybee & 48.7/50.6 & 50.8/53.1 & 42.7/44.8 & 48.7/50.3 & 49.2/47.1 & 49.7/53.4 & 40.8/43.7 & 42.7/44.5 & 51.3/54.2 & 48.5/46.1 & 48.2/51.9 \\
GAP + Pargo & 36.5/33.1 & 28.7/25.4 & 34.6/37.5 & 28.3/31.5 & 44.6/47.4 & 31.2/28.4 & 23.3/19.9 & 22.8/20.6 & 28.7/25.0 & 29.9/29.4 & 33.2/35.1 \\
\textbf{MVX-LLM (Ours)} & 59.6/{\color[HTML]{3531FF} \textbf{61.0}} & {\color[HTML]{3531FF} \textbf{53.5}}/{\color[HTML]{3531FF} \textbf{55.7}} & 51.7/53.6 & {\color[HTML]{3531FF} \textbf{55.2}}/{\color[HTML]{3531FF} \textbf{58.1}} & {\color[HTML]{3531FF} \textbf{58.3}}/{\color[HTML]{3531FF} \textbf{61.2}} & {\color[HTML]{3531FF} \textbf{55.2}}/{\color[HTML]{3531FF} \textbf{58.2}} & {\color[HTML]{3531FF} \textbf{51.3}}/{\color[HTML]{3531FF} \textbf{53.9}} & {\color[HTML]{3531FF} \textbf{57.3}}/{\color[HTML]{3531FF} \textbf{59.4}} & 57.4/60.3 & 52.8/55.6 & {\color[HTML]{3531FF} \textbf{55.5}}/{\color[HTML]{3531FF} \textbf{58.1}} \\ 
\bottomrule
\end{tabular}%
}
\end{table*}
Table~\ref{tab:main_results} presents comprehensive comparisons across different architectural approaches and fusion strategies on the \textsc{DriveXQA} dataset. We evaluate both projector architectures and attention-based fusion mechanisms under various environmental conditions and sensor failure scenarios. Human evaluation on a subset of samples shows strong correlation with GPT-4o-mini scores across all methods and conditions, with Spearman correlation coefficients~\cite{majumdar2024openeqa} consistently above $0.83$, confirming the reliability of our automated evaluation approach. \textsc{MVX-LLM} framework with DCA mechanism significantly outperforms both token fusion methods and traditional projector architectures. Specifically, our full system (Ours) achieves the highest average GPTScore of $55.5$, surpassing the best token fusion method QAttn(Spectral) ($52.8$) by $3\%$ and the strongest projector architecture GAP + Honeybee ($48.2$) by $7.3\%$.
Notably, our approach shows remarkable improvements in challenging conditions such as foggy ($53.5$ \textit{vs.} $25.1$ for prepend, $53.7$ for QAttn(Spectral) token fusion, and $50.8$ for GAP + Honeybee projector) and various sensor failure scenarios. The consistent performance gains across all environmental conditions and sensor degradation types highlight the effectiveness of our DCA mechanism and QAttn for robust multi-modal fusion, demonstrating clear advantages over both conventional token-level fusion strategies and projector-based integration approaches.
\subsection{Ablation Study}
\subsubsection{Attention Component Ablation Analysis}
Table~\ref{tab:dca_ablation} examines the contributions of spatial cross-attention (sAttn), channel cross-attention (cAttn), and Query Attention (QAttn) for multi-modal fusion. Individual mechanisms show limited performance, with channel attention alone achieving only (GPTScore: $20.4$) and spatial attention alone reaching (GPTScore: $24.0$), indicating that neither mechanism is sufficient for effective multi-modal fusion independently.
Dual attention combinations demonstrate significant improvements. The cAttn+QAttn combination achieves (GPTScore: $34.8$), while sAttn+QAttn reaches (GPTScore: $41.1$), showing that spatial attention provides more effective cross-modal alignment than channel attention when combined with learnable query mechanisms. 
The sAttn+cAttn combination performs best among dual mechanisms (GPTScore: $50.5$), verifying the superiority of our dual cross-attention design.
The complete triple attention mechanism (sAttn+cAttn+QAttn) achieves optimal performance (GPTScore: $55.5$), demonstrating the complementary effects of all components: spatial cross-attention enables position-level cross-modal alignment, channel cross-attention facilitates semantic information exchange, and QAttn provides adaptive spatial attention for task-relevant focus in autonomous driving scenarios.
\begin{table}[h]
\centering
\caption{Ablation study on attention mechanism components. Results demonstrate the individual and combined contributions of spatial cross-attention (sAttn), channel cross-attention (cAttn), and Query Attention (QAttn) for robust multi-modal fusion.}
\label{tab:dca_ablation}
\setlength{\tabcolsep}{1mm}
\resizebox{\columnwidth}{!}{%
\begin{tabular}{ccc|cccccc}
\toprule
\textbf{sAttn}& \textbf{cAttn}& \textbf{QAttn}&\textbf{CIDEr} & \textbf{BLEU-4} & \textbf{ROUGE-L} & \textbf{METEOR} & \textbf{Sim} & \textbf{GPTScore} \\ \midrule
\cellcolor{green!20}$\checkmark$ & \cellcolor{gray!15} & \cellcolor{gray!15} & 48.3 & 9.4 & 23.5 & 29.8 & 49.1 & 24.0 \\
\cellcolor{green!20}$\checkmark$ & \cellcolor{gray!15} & \cellcolor{green!20}$\checkmark$ & 50.6 & 15.7 & 33.2 & 41.4 & 61.1 & 41.1 \\
\cellcolor{gray!15} & \cellcolor{green!20}$\checkmark$ & \cellcolor{gray!15} & 31.4 & 6.5 & 19.7 & 26.6 & 46.8 & 20.4 \\
\cellcolor{gray!15} & \cellcolor{green!20}$\checkmark$ & \cellcolor{green!20}$\checkmark$ & 42.9 & 11.3 & 27.0 & 35.6 & 57.0 & 34.8 \\
\cellcolor{green!20}$\checkmark$ & \cellcolor{green!20}$\checkmark$ & \cellcolor{gray!15} & 105.7 & 24.6 & 46.0 & 56.6 & 70.5 & 50.5\\
\cellcolor{green!20}$\checkmark$ & \cellcolor{green!20}$\checkmark$ & \cellcolor{green!20}$\checkmark$ & {\color[HTML]{3531FF} \textbf{144.6}} & {\color[HTML]{3531FF} \textbf{28.1}} & {\color[HTML]{3531FF} \textbf{50.6}} & {\color[HTML]{3531FF} \textbf{61.2}} & {\color[HTML]{3531FF} \textbf{74.6}} & {\color[HTML]{3531FF} \textbf{55.5}} \\ 
\bottomrule
\end{tabular}%
}
\end{table}
\begin{table*}[!t]
\vspace{2mm}
\centering
\caption{Ablation study on sensor modality combinations. Results show the importance of each modality and its complementary effects for robust VQA performance under adverse conditions.}
\label{tab:ablation}
\resizebox{\textwidth}{!}{%
\setlength{\tabcolsep}{4mm}
\setlength{\arrayrulewidth}{0.5pt}
\begin{tabular}{cccc|cccccc}  
\toprule
 \textbf{RGB} & \textbf{Depth} & \textbf{Event} & \textbf{LiDAR} & \textbf{CIDEr} & \textbf{BLEU-4} & \textbf{ROUGE-L} & \textbf{METEOR} & \textbf{Sim} & \textbf{GPTScore}\\ \midrule
\cellcolor{green!20}$\checkmark$ & \cellcolor{green!20}$\checkmark$ & \cellcolor{gray!15} & \cellcolor{gray!15} & {\color[HTML]{3531FF} \textbf{163.4}} & 19.5 & 31.6 & 38.1 & 53.4 & 23.4 \\
\cellcolor{green!20}$\checkmark$ & \cellcolor{gray!15} & \cellcolor{green!20}$\checkmark$ & \cellcolor{gray!15} & 152.6 & 26.1 & 43.4 & 48.2 & 63.1 & 42.4 \\
\cellcolor{gray!15} & \cellcolor{green!20}$\checkmark$ & \cellcolor{green!20}$\checkmark$ & \cellcolor{gray!15} & 8.1 & 3.9 & 16.9 & 23.0 & 43.5 & 18.9 \\
\cellcolor{green!20}$\checkmark$ & \cellcolor{green!20}$\checkmark$ & \cellcolor{green!20}$\checkmark$ & \cellcolor{gray!15} & 157.6 & 27.3 & 46.2 & 51.7 & 68.7 & 46.0 \\
\cellcolor{green!20}$\checkmark$ & \cellcolor{green!20}$\checkmark$ & \cellcolor{gray!15} & \cellcolor{green!20}$\checkmark$ & 117.5 & 26.0 & 46.7 & 54.1 & 71.5 & 50.7\\
\cellcolor{green!20}$\checkmark$ & \cellcolor{gray!15} & \cellcolor{green!20}$\checkmark$ & \cellcolor{green!20}$\checkmark$ & 50.1 & 13.4 & 30.0 & 37.6 & 57.3 & 34.2 \\
\cellcolor{gray!15} & \cellcolor{green!20}$\checkmark$ & \cellcolor{green!20}$\checkmark$ & \cellcolor{green!20}$\checkmark$ & 77.4 & 17.7 & 35.3 & 43.2 & 62.9 & 39.7 \\
\cellcolor{green!20}$\checkmark$ & \cellcolor{green!20}$\checkmark$ & \cellcolor{green!20}$\checkmark$ & \cellcolor{green!20}$\checkmark$ & 144.6 & {\color[HTML]{3531FF} \textbf{28.1}} & {\color[HTML]{3531FF} \textbf{50.6}} & {\color[HTML]{3531FF} \textbf{61.2}} & {\color[HTML]{3531FF} \textbf{74.6}} & {\color[HTML]{3531FF} \textbf{55.5}} \\ 
\bottomrule
\end{tabular}%
}
\end{table*}
\subsubsection{Multi-Modal Sensor Combination Analysis}
To systematically evaluate the contribution of different sensor modality combinations in \textsc{MVX-LLM} framework, we conducted a comprehensive ablation study using the QAttn token aggregation mechanism. We examined various multi-modal combinations to understand their complementary effects and potential interference patterns. 
Table~\ref{tab:ablation} presents the detailed results across multiple evaluation metrics.
Two-modality combinations reveal distinct patterns in sensor complementarity. 
RGB+Depth achieves moderate performance (GPTScore: $23.4$), while RGB+Event demonstrates stronger complementary effects (GPTScore: $42.4$). 
Notably, the Depth+Event combination without RGB guidance performs poorly (GPTScore: $18.9$), highlighting the fundamental importance of RGB visual information as the foundation for multi-modal fusion in autonomous driving scenarios.
Three-modality combinations show more complex interaction dynamics. 
RGB+Depth+Event achieves substantial improvement (GPTScore: $46.0$), indicating that when properly grounded with RGB, multiple visual modalities can work synergistically. RGB+Depth+LiDAR (GPTScore: $50.7$) demonstrates the strongest three-modality performance through effective integration of visual and geometric information, while RGB+Event+LiDAR (GPTScore: $34.2$) shows more modest gains. Interestingly, Depth+Event+LiDAR without RGB guidance achieves moderate performance (GPTScore: $39.7$), suggesting that geometric information from LiDAR can partially compensate for the absence of RGB visual context, though not as effectively as RGB-grounded combinations.  Our complete four-modality MVX-LLM system achieves optimal performance (GPTScore: $55.5$), verifying that the QAttn mechanism effectively leverages all complementary sensor information while maintaining robust fusion across diverse modalities.

\subsection{Qualitative Analysis}
\begin{figure}[h]
    \centering
    \includegraphics[width=0.95\columnwidth]{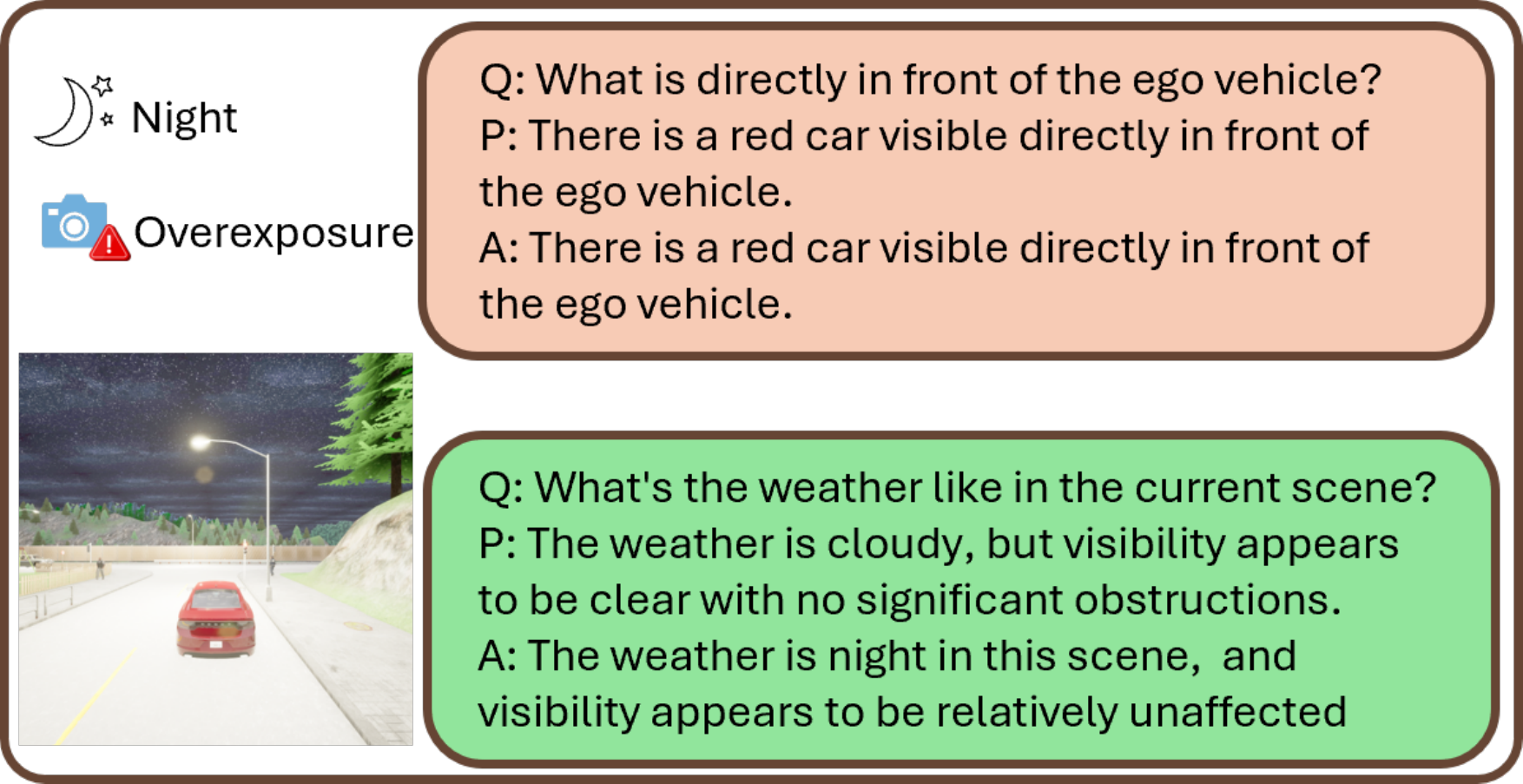}
    \caption{Qualitative analysis of multi-modal fusion performance under night conditions with camera overexposure.}
    \label{fig:case_analysis}
\end{figure}

To further illustrate the effectiveness of our multi-modal fusion approach, we analyze specific prediction cases that demonstrate the complementary nature of different sensor modalities under adverse conditions. 
Fig.~\ref{fig:case_analysis} presents a representative example from the \textsc{DriveXQA} dataset showing the model's performance under combined night conditions and camera overexposure.
In the upper case, when asked about objects directly in front of the ego vehicle, our model correctly identifies the red car despite the challenging lighting and overexposure conditions. This success demonstrates the strength of our multi-modal fusion: while the RGB camera suffers from overexposure artifacts, complementary modalities such as LiDAR provide robust geometric information about vehicle positions and shapes. The DCA mechanism appears to leverage depth and point cloud data for object detection even when visual sensors are compromised, consistent with strong RGB+Depth+LiDAR performance (GPTScore: $50.7$) in our quantitative results. Conversely, the weather recognition error suggests potential limitations when geometric modalities must compensate for compromised visual cues. This aligns with our ablation findings showing performance variations across different modality combinations.
This analysis reveals that geometric sensors effectively compensate for object detection tasks, but semantic environmental understanding remains critically dependent on functional visual sensors.
\subsection{Implications for Sim-to-Real Transfer}
Despite the sim-to-real gap, our benchmark provides a controlled stress-test environment by systematically modeling adverse weather and sensor degradations (\textit{e.g.}, motion blur, over/under-exposure, LiDAR jitter, and low-resolution event streams). This setup mirrors common deployment-time failures and exposes multimodal fusion bottlenecks under safety-critical conditions. As a result, it supports more reliable method comparison and helps validate robustness improvements before transitioning from simulation to real-world autonomous driving.
\section{Conclusion}
In this work, we introduce the \textsc{DriveXQA} dataset and \textsc{MVX-LLM} framework for multi-modal autonomous driving scene understanding under adverse conditions. Our dataset evaluates three semantic levels: global environmental assessment, allocentric spatial reasoning, and ego-vehicle operational analysis. \textsc{MVX-LLM} addresses the limitation that existing MLLMs cannot process multiple complementary modalities by introducing a Dual Cross-Attention mechanism that fuses RGB, depth, event camera, and LiDAR data. Our method achieves GPTScore $55.5$, demonstrating effective multi-modal fusion for challenging driving scenarios. 
\subsection{Limitations and Future Work} 
While our approach demonstrates promising results, several limitations remain. First, \textsc{DRIVEXQA} is built on CARLA simulation data; while this enables systematic control over adverse conditions, physical phenomena such as LiDAR scattering in heavy rain or DVS noise in low-light are only approximated in simulation, and validation on real-world corrupted sensor data remains important future work. Second, the benchmark is limited to single-frame understanding, treating scenes independently without temporal context; we adopt this static formulation as a controlled baseline to isolate cross-modal spatial reasoning, with video-based temporal extension planned as direct future work. Third, our modality-combination ablations provide evidence for cross-modal complementarity under adverse conditions; explicit sensor-dropout evaluation at inference time remains a valuable direction to further substantiate compensation dynamics. Future work will incorporate real-world data, temporal reasoning, and lightweight onboard deployment.

\clearpage
\section*{Acknowledgement}
This work was supported in part by National Natural Science Foundation of China under Grant No. 62503166 and No. 62473139, in part by the Hunan Provincial Research and Development Project (Grant No. 2025QK3019), in part by the State Key Laboratory of Autonomous Intelligent Unmanned Systems (the opening project number ZZKF2025-2-10), in part by the Deutsche Forschungsgemeinschaft (DFG, German Research Foundation) - SFB 1574 - 471687386, and in part by Helmholtz Association of German Research Centers, in part by the Ministry of Science, Research and the Arts of Baden-W\"urttemberg (MWK) through the Cooperative Graduate School Accessibility through AI-based Assistive Technology (KATE) under Grant BW6-03, and in part by the Helmholtz Association Initiative and Networking Fund on the HAICORE@KIT and HOREKA@KIT partition. This research was partially funded by the Ministry of Education and Science of Bulgaria (support for INSAIT, part of the Bulgarian National Roadmap for Research Infrastructure).

{
    \small
    \bibliographystyle{ieeenat_fullname}
    \bibliography{main}
}
\end{document}